\documentclass[conference]{IEEEtran}
\IEEEoverridecommandlockouts
\usepackage{cite}
\usepackage{amsmath,amssymb,amsfonts}
\usepackage{algorithmic}
\usepackage{graphicx}
\usepackage{textcomp}
\usepackage{xcolor}
\usepackage{caption}
\usepackage{bm}
\usepackage[labelformat=simple]{subcaption}
\usepackage[a4paper, total={184mm,239mm}]{geometry}
\def\BibTeX{{\rm B\kern-.05em{\sc i\kern-.025em b}\kern-.08em
    T\kern-.1667em\lower.7ex\hbox{E}\kern-.125emX}}

\usepackage{censor}
\StopCensoring

\begin{document}

\title{LLM-based AI Agent for Sizing of Analog and Mixed Signal Circuit\\

\thanks{\blackout{Chang Liu is sponsored by Peter Denyer's Scholarship at The University of Edinburgh}}

}

\author{
\IEEEauthorblockN{\blackout{Chang Liu}}
\IEEEauthorblockA{\textit{\blackout{School of Engineering}} \\
\textit{\blackout{The University of Edinburgh}}\\
\blackout{Edinburgh, UK} \\
\blackout{s2598777@ed.ac.uk}}
\and
\IEEEauthorblockN{\blackout{Emmanuel A. Olowe}}
\IEEEauthorblockA{\textit{\blackout{School of Engineering}} \\
\textit{\blackout{The University of Edinburgh}}\\
\blackout{Edinburgh, UK} \\
\blackout{e.a.olowe@sms.ed.ac.uk}}
\and
\IEEEauthorblockN{\blackout{Danial Chitnis}}
\IEEEauthorblockA{\textit{\blackout{School of Engineering}} \\
\textit{\blackout{The University of Edinburgh}}\\
\blackout{Edinburgh, UK} \\
\blackout{d.chitnis@ed.ac.uk}}
}

\maketitle

\begin{abstract}
The design of Analog and Mixed-Signal (AMS) integrated circuits (ICs) often involves significant manual effort, especially during the transistor sizing process. While Machine Learning techniques in Electronic Design Automation (EDA) have shown promise in reducing complexity and minimizing human intervention, they still face challenges such as numerous iterations and a lack of knowledge about AMS circuit design. Recently, Large Language Models (LLMs) have demonstrated significant potential across various fields, showing a certain level of knowledge in circuit design and indicating their potential to automate the transistor sizing process. In this work, we propose an LLM-based AI agent for AMS circuit design to assist in the sizing process. By integrating LLMs with external circuit simulation tools and data analysis functions and employing prompt engineering strategies, the agent successfully optimized multiple circuits to achieve target performance metrics. 
We evaluated the performance of different LLMs to assess their applicability and optimization effectiveness across seven basic circuits, and selected the best-performing model Claude~3.5 Sonnet for further exploration on an operational amplifier, with complementary input stage and class AB output stage. This circuit was evaluated against nine performance metrics, and we conducted experiments under three distinct performance requirement groups. A success rate of up to 60\% was achieved for reaching the target requirements. Overall, this work demonstrates the potential of LLMs to improve AMS circuit design. 
\end{abstract}

\begin{IEEEkeywords}
Analogue and Mixed-Signal (AMS), Large Language Models (LLMs), Prompt Engineering.
\end{IEEEkeywords}

\section{Introduction}
\begin{figure}[t]
    \centering
    \includegraphics[width=0.45\textwidth]{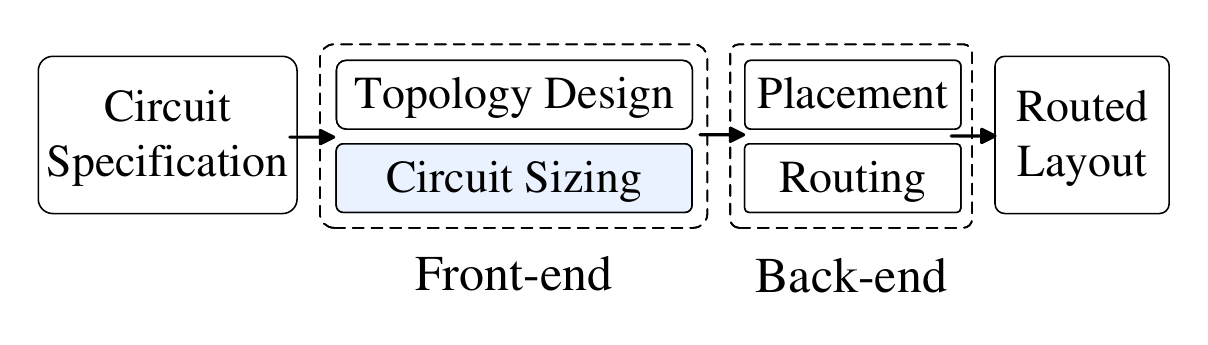}
    \caption{Typical circuit design flow}
    \label{fig:design_flow}
\end{figure}

Analog and mixed-signal (AMS) integrated circuits, such as Analog-to-Digital Converters (ADCs) \cite{walden1999analog}, filters \cite{winder2002analog}, and Power Management Integrated Circuits (PMICs) \cite{ballo2021review}, play a vital role in communication systems and consumer electronics. The design process of AMS circuits typically follows a structured flow, beginning with the front-end stages of topology selection and circuit sizing, followed by the back-end tasks of placement and routing, as illustrated in Fig.~\ref{fig:design_flow}. Among these stages, circuit sizing remains particularly challenging due to its high-dimensional design space and complex trade-offs between performance metrics. This complexity often demands significant human expertise and iterative fine-tuning. Therefore, as a crucial aspect of Electronic Design Automation (EDA), automatic circuit sizing has garnered increasing research interest. \cite{lyu2018batch}, \cite{liao2017parasitic}.

\begin{figure*}[t]
\centering
\includegraphics[width=1\textwidth]{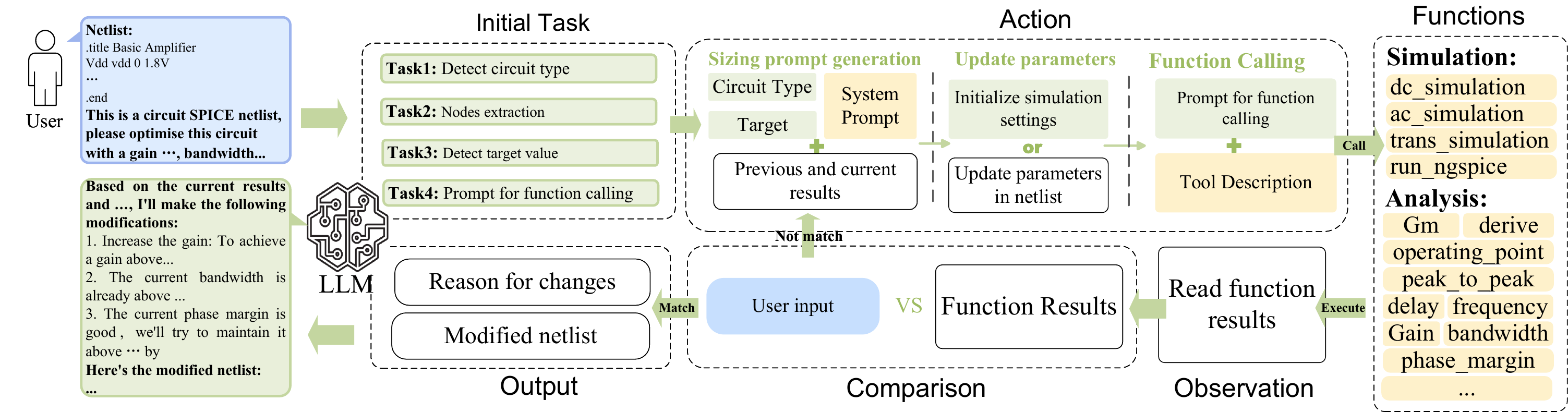}
\caption{The entire process for circuit sizing with the proposed AI Agent. The process begins with the task decomposition stage, generating four tasks for different stages. Action, observation and comparison formed a ReAct optimization loop. Finally, the agent generates output consisting of reasons for changes and modifications of the netlist for the user.}
\label{fig:whole process}
\end{figure*}

The rapid expansion of Machine Learning (ML) techniques has increasingly highlighted its potential for automating circuit design, offering new possibilities for achieving automation in the circuit sizing process. Bayesian Optimization (BO) models transistor sizing as a black-box optimization problem, employing Gaussian Process (GP) regression to efficiently guide the search for optimal solutions \cite{lyu2018batch, touloupas2021local}. Deep Reinforcement Learning (DRL) further improves sizing efficiency by leveraging rewards from iterative simulations, enabling adaptation to complex design spaces \cite{zhao2022analog}. However, these methods operate without domain-specific analog design knowledge, leading to high computational costs and inconsistent performance across varying circuit designs and performance metrics. \cite{settaluri2020autockt}.

Recently, large language models (LLMs) such as Claude~3.5 \cite{TheC3}, GPT-4 \cite{achiam2023gpt}, and Llama~3 \cite{touvron2023llama} have demonstrated exceptional capabilities in tasks such as natural language processing, code generation, and reasoning. Several studies have explored the use of LLMs to enable automated schematic design for analog circuits. However, few of them focus on sizing. For example, Artisan leverages domain-specific LLMs to automate operational amplifier netlist generation and utilize gm/Id method for transistor sizing \cite{chen2024artisan}. Another tool, AmpAgent utilizes Retrieval Augmented Generation (RAG) to address LLM's knowledge gaps in multi-stage amplifiers and use conventional algorithms to size the device \cite{liu2024ampagent}. LEDRO combined LLMs with optimization techniques TuRBO to iteratively reduce the design space \cite{kochar2024ledro}. While these approaches achieve higher success rates and shorten execution time compared to conventional algorithms, their scope remains limited to basic performance metrics, which are insufficient for addressing the complex and diverse requirements of modern AMS circuit design. 

In this paper, we propose an LLM-based AI agent to optimize the transistor sizing process in AMS circuit design. Our approach integrates large language models (LLMs) with the Ngspice simulator and custom data analysis functions, enabling automated in-loop simulations that significantly reduce manual intervention. We evaluate the performance of various LLM APIs across seven basic circuits to assess their applicability in the field of circuit design. Based on this comparison, we select the top-performing model, Claude 3.5 Sonnet, to explore more complex circuit configurations with stricter performance metrics.

%By leveraging the in-context learning and reasoning capabilities of LLMs, we employ prompt engineering to provide the agent with structured feedback, including current and simulation results history, netlists, and design parameters. This feedback mechanism allows the agent to iteratively learn from previous design iterations, refine transistor sizing decisions, and enhance overall circuit performance. The feedback structure is pre-defined within the agent. 

\section{Methodology}
The proposed agent takes a circuit netlist and expected performance metrics as user inputs and ultimately generates an optimized netlist that meets the target performance, along with the reasons for parameter adjustments.

\subsection{LLM-Based AI Agent}
The entire process for circuit sizing with the proposed AI Agent begins with task decomposition based on user input. In the optimization process, action, observation, and comparison create a Reasoning and Acting (ReAct) loop \cite{yao2022react}, where the agent cycles through reasoning (analyzing) and taking action based on its insights. It starts with sizing prompt generation. The agent then selects appropriate functions based on performance metrics. After executing these functions, it reviews the results and compares them with the user’s input to check for alignment. The final output is produced if the results meet the performance criteria. To improve the agent's interpretability, this agent will provide not only the modified netlist but also the reasoning behind each parameter adjustment in every iteration. The entire process for circuit sizing with the proposed AI Agent is shown in Fig.~\ref{fig:whole process}.

\subsection{Prompt Engineering}
For optimizing AMS circuits, prompt engineering enables the customization of LLM responses to meet specific design requirements through carefully crafted instructions. In this work, we employ Chain-of-Thought (CoT), a technique that enhances the model’s decision-making process by generating intermediate reasoning steps \cite{wei2022chain}. This approach streamlines the transistor sizing process and ensures more efficient and accurate design optimization.

%\begin{figure}[t]
%    \centering
%    \includegraphics[width=0.5\textwidth]{tem.pdf}
%    \caption{Prompt template for optimization. The template is pre-defined by the %developer. Four main components are passed as parameters to complete the prompt, which %is suitable for all the circuits under test.}
%    \label{fig:tem}
%\end{figure}

To achieve a COT, we developed a prompt template that consisted of a system prompt, four parameters, and CoT instructions. The system prompt includes constraints such as maintaining the power supply voltage and using the same transistor model. The four main parameters: circuit type, previous results, current results and target performance provide highly relevant context for LLMs, enabling effective in-context learning, which helps LLMs better understand the relationship between performance and parameters. The CoT instructions guide the LLMs to think step by step based on their observation to prompt them to generate new, potentially optimal design points. These points are then simulated, and the results are used to further enrich the context history. 

\subsection{Simulation in-loop}
We set up the simulation in-loop by function calling. The functions used for optimization are complex and highly interconnected, creating a workflow where each step relies on the outputs of the previous one, ensuring a seamless optimization process. Therefore, we utilized CoT to guide the LLM thinking step by step to choose required functions like a human AMS circuit designer, taking into account the entire optimization flow, including selecting the appropriate simulation type based on the required performance metrics, running simulations tool, and analyzing the performance with the corresponding functions. All the functions are pre-defined in the agent by the developer.

\section{Results}

\subsection{The Performance of Different LLMs}\label{AA}
Initially, We evaluate the performance of five LLM APIs — Claude~3.5 Sonnet, Claude~3 Sonnet, Claude~3 Haiku, \mbox{GPT-4o}, and \mbox{GPT-4o} mini — by optimizing seven different types of circuits. The assessment uses a consistent set of performance metrics and focuses on the number of iterations each LLM needs to successfully optimize a circuit, with a maximum limit of 20 iterations.

\begin{figure*}[t]
\centerline{\includegraphics[width=1\textwidth]{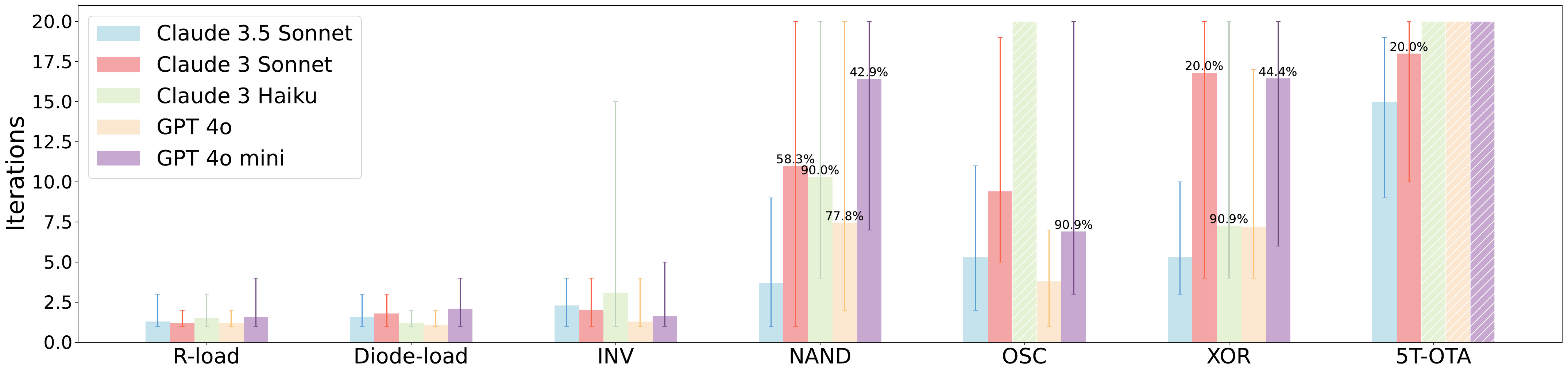}}
\caption{Performance comparison of different LLMs. This chart compares the performance of various LLMs across different circuits within 20 iterations. The height of the bars represents the average number of iterations, with lower values indicating better optimization performance. Error bars show the range of iterations observed across ten attempts. In each attempt, exceeding 20 iterations is considered a failure. The success rate smaller than 100\% is labelled on the bar. Hatched bars indicate that the model failed to optimize specific circuits in all ten attempts. All the tests were performed in September of 2024.}
\label{fig: performance}
\end{figure*}

The results are shown in Fig.~\ref{fig: performance}. Claude 3.5 Sonnet outperformed other LLMs, achieving the highest success rate, the fewest iterations, and the lowest variation, demonstrating strong stability in circuit sizing. \mbox{GPT-4o} showed comparable performance but failed on the \mbox{5T-OTA}. Claude 3 Sonnet and \mbox{GPT-4o} mini exhibited similar results, with lower success rates and higher variability, especially for NAND and XOR gates. Claude 3 Haiku performed well on digital circuits but failed on the oscillator and \mbox{5T-OTA}. Therefore, we use Claude 3.5 Sonnet for further exploration in more complex circuit design tasks.

\subsection{An Operational Amplifier}

\begin{figure}[t]
    \centering
    \resizebox{0.5\textwidth}{!}{%
        \includegraphics{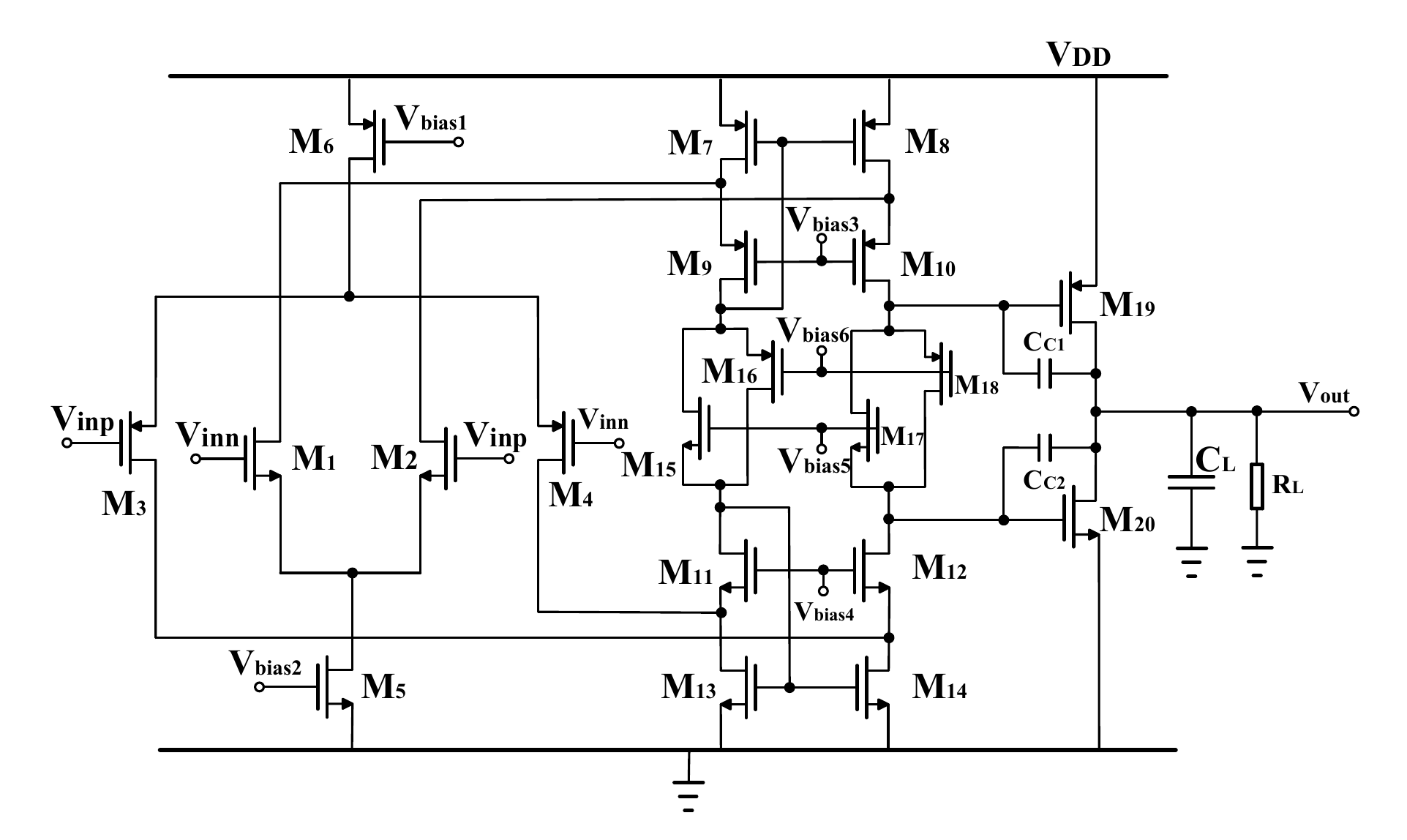}
    }
    \caption{Schematic of the tested opamp, featuring a complementary input stage and a class AB output stage.
}
    \label{fig:schematic}
\end{figure}

The agent is employed to optimize an opamp with a complementary input stage and class AB output stage, which is consist of 20 transistors \cite{hogervorst1994compact}. The schematic is shown in Fig.~\ref{fig:schematic}. 

\begin{table}[t!]
\centering
\renewcommand{\arraystretch}{1.2}
\caption{Performance evaluation for 5 attempts across three groups (G1, G2 and G3). A 5\% tolerance is applied to all metrics; deviations beyond this tolerance are marked in \textcolor{red}{red}. Different targets for G1 and G2, G1 and G3 are marked in \textbf{bold}. If the 25 iteration is reached without meeting any target, it is marked as a failure.}
\resizebox{0.5\textwidth}{!}{ % Adjust width as needed
\begin{tabular}{c|c|c|c|c|c|c|c|c|c|c}
\hline
\rotatebox{90}{\textbf{Group}} & \rotatebox{90}{\textbf{Gain  (dB)}} & \rotatebox{90}{\textbf{UGBW  (MHz)}} & \rotatebox{90}{\textbf{PM ($^\circ$)}} & \rotatebox{90}{\textbf{Power  (mW)}} & \rotatebox{90}{\textbf{CMRR  (dB)}} & \rotatebox{90}{\textbf{THD (dB)}} & \rotatebox{90}{\textbf{CL (F)  RL($\Omega$)}} & \rotatebox{90}{\textbf{Offset  (V)}} & \rotatebox{90}{\textbf{Output Range (V)}} & \rotatebox{90}{\textbf{Iter}}\\
\hline
\hline
\textbf{G1 Target} &$\geq65$ & $\geq10$ & $\geq55$ & $\leq10$ & $\geq100$& $\leq-26$ & \text{10P, 1k}& $\leq1$ & $\geq1.75$ & 25 \\
\hline
G1-1 & 67.91 & 19.95 & 61.59 & 4.8 & 110.76 & -26.06 & \text{10P, 1k} & 0.98 & 1.68 & \textbf{13}   \\
G1-2 & 68.63 & 19.95 & 72.26 & \textcolor{red}{13.22} & 97.63 & -26.09 & \text{10P, 1k} & \textcolor{red}{5.30} & \textcolor{red}{1.24} & fail   \\
G1-3 & 68.57 & 15.85 & 71.75 & 4.53 & 118.61 & -26.02 & \text{10P, 1k} & 0.16 & \textcolor{red}{1.32} & fail   \\
G1-4 & 66.07 & 12.58 & 54.50 & 7.69 & 124.13 & -25.42 & \text{10P, 1k} & 0.40 & 1.68 & \textbf{20}   \\
G1-5 & 66.45 & 50.12 & 60.59 & 7.43 & 105.62 & -26.24 & \text{10P, 1k} & 0.69 & 1.67 & \textbf{23}  \\
\hline
\hline
G2 Target & $\geq65$ & \bm{$\geq 5$} & \bm{$\geq 45$} & \bm{$\leq 5$} & $\geq100$ & $\leq-26$ & \textbf{\text{50P, 100k}} & $\leq1$ & $\geq1.75$ & 25 \\\hline
G2-1 & \textcolor{red}{58.73} & 5.01 & 78.98 & 4.2 & 100.89 & -25.89 & \text{50P, 100k} & \textcolor{red}{1.80} & 1.67 & fail   \\
G2-2 & 66.76 & 12.59 & \textcolor{red}{34.14} & 1.2 & 131.64 & -24.75 & \text{50P, 100k} & \textcolor{red}{4.50} & 1.77 & fail   \\
G2-3 & 63.27 & 9.99 & 52.34 & 1.5 & 111.34 & -25.98 & \text{50P, 100k} & 0.99 & 1.69 & \textbf{16}   \\
G2-4 & 62.19 & 7.94 & 55.19 & 0.7 & 131.19 & -26.29 & \text{50P, 100k} & 0.42 & 1.70 & \textbf{25}  \\
G2-5 & 66.74 & 15.85 & 63.44 & 4.9 & 112.01 & -26.29 & \text{50P, 100k} & 0.86 & 1.70 & \textbf{24}   \\\hline
\hline
G3 Target & $\geq65$ & \bm{$\geq50$} & $\geq55$ & \bm{$\leq50$} & $\geq100$ & $\leq-26$ & 10P, 1k & \bm{$\leq5$} & \bm{$\geq1.7$} & 25 \\\hline
G3-1 & \textcolor{red}{50.97} & 99.99 & 56.04 & 15.83 & \textcolor{red}{93.70} & -30.16 & 10P, 1k & 5.16 & \textcolor{red}{1.66} & fail   \\
G3-2 & 68.91 & 63.09 & 64.55 & 20.27 & 96.77 & -24.68 & 10P, 1k & 4.30 & \textcolor{red}{1.39} & fail   \\
G3-3 & \textcolor{red}{58.97} & 50.12 & 73.84 & 4.12 & 95.74 & -41.57 & 10P, 1k & \textcolor{red}{42.25} & \textcolor{red}{0.68} & fail   \\
G3-4 & 69.33 & 63.09 & 67.82 & 10.08 & 108.78 & -27.22 & 10P, 1k & 2.70  & 1.65 & \textbf{11}  \\
G3-5 & 68.38 & 79.43 & 51.21 & 47.01 & \textcolor{red}{81.73} & -32.62 & 10P, 1k  & \textcolor{red}{14.27} & 1.69 & fail  \\
\hline

\end{tabular}}
\label{tab:performance_metrics}
\end{table}

 We consider nine performance metrics of opamps in three configuration groups. These three groups, named G1, G2, and G3 differ in bandwidth, output load, and power. In all groups, the supply voltage is 1.8~V, and the transistor model is PTM 180~nm \cite{cao2011predictive}. Table.~\ref{tab:performance_metrics} shows the performance metrics of each group, including the passed and failed attempts and success rate among five independent trials.

\begin{figure}[t]
    \centering
    \resizebox{0.5\textwidth}{!}{%
        \includegraphics{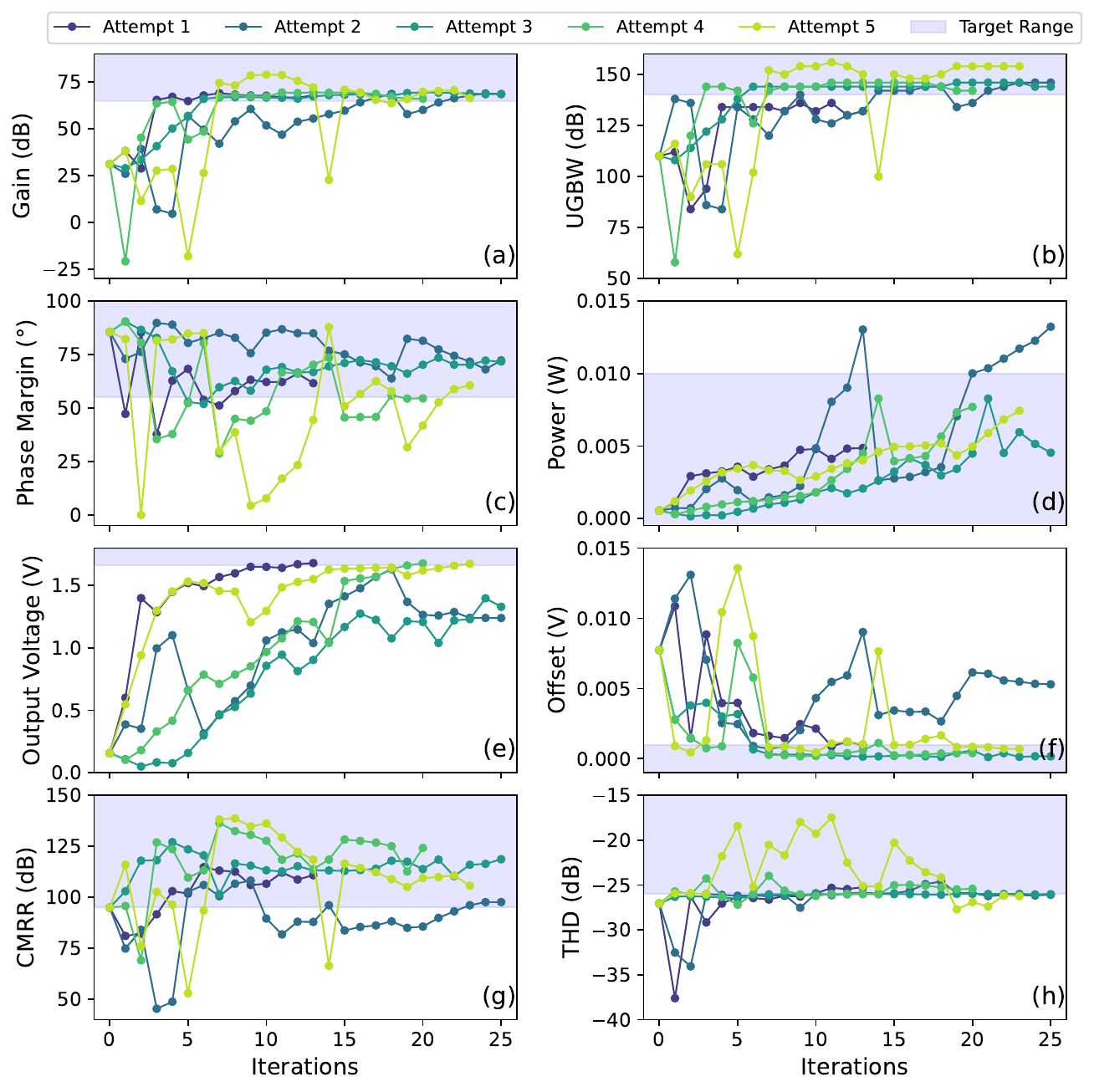}
    }
    \caption{Optimization results for the opamp. (a) Gain (b) Unity-Gain Bandwidth (c) Phase Margin (d) Power (e) Input Offset (f) Output Voltage Range (g) CMRR (h) THD. (b) and (e) are tested under unity gain configuration, others are tested open loop. (a), (b), (c), (d) and (g) are tested at Vin,cm = 0.9~V, (e), (f), (h) are tested across the Vin from 0 to 1.8~V. Those already in the target range initially or after a few iterations may still fluctuate or even get worse due to the optimization process prioritizing other performance metrics. This occurs because the agent has not yet achieved a balance between all required performance criteria.}
    \label{fig:4*2}
\end{figure}

 The optimization process of the opamp under G1 is illustrated in Fig.~\ref{fig:4*2}. Fig. ~\ref{fig:4*2} (e) and  Fig. ~\ref{fig:4*2} (f) underscore the challenges in achieving low input offset voltage and high output voltage range targets. Power consumption increases due to the high drive requirements for wide output swing. Despite fluctuations in all the figures, the agent succeeds in 3 out of 5 attempts within 25 iterations, proving the effectiveness of in-context learning and our prompt strategy.

 %[input offset, output voltage range]

For G2, targeting low power and a higher load capacitor, unity-gain bandwidth and phase margin constraints were relaxed to account for trade-offs with the load. The agent achieved a 60\% success rate within 25 iterations. For G3, it prioritized a higher bandwidth and power relative to G1 and G2. The transistor sizes and bias voltages belonging to a successful attempt of each configuration group are shown in Table~\ref{tab:transistor_sizing}.

\begin{table}[h!]
\centering
\caption{Transistor size and bias voltages for G1-5, G2-4 and G3-4 after optimization}
\resizebox{0.5\textwidth}{!}{
\begin{tabular}{c|c|c|c}
\hline
\textbf{Transistor} & \textbf{G1-5 (W/L, \text{µm}/\text{µm})} & \textbf{G2-4 (W/L, \text{µm}/\text{µm})} &  \textbf{G3-4 (W/L, \text{µm}/\text{µm})} \\ \hline
M1, M2 & 53/1.85 & 65/0.48 & 23/0.2 \\ 
M3, M4 & 53/1.85 &32/0.48 & 11.5/0.2\\ 
M5 & 9/0.45 & 2.5/0.46 & 15/0.6\\ 
M6 & 9/0.45 & 5/0.46 & 15/0.6 \\ 
M7, M8 & 80/1.8 & 25.5/0.46 & 20/0.22\\ 
M9, M10 & 80/1.8 & 25.5/0.46 & 20/0.22\\ 
M11, M12 & 6/0.45 & 4.3/0.42 & 8/0.2\\ 
M13, M14 & 5/0.45 & 3.7/0.385 &  8/0.2 \\ 
M15, M17 & 15/0.9 & 5.8/0.51 & 11/0.18\\
M16, M18 & 30/0.9 & 11.6/0.51 & 22/0.18\\ 
M19 &340/0.43 & 90/0.53 & 100/0.18 \\ 
M20 &170/0.43 & 45/0.53 & 50/0.18 \\ 
\hline
\hline
bias1 & 0.88 & 0.89 & 0.75 \\
bias2 & 1.05 & 0.95 & 1.1 \\
bias3 & 0.46 & 0.71 & 1.13 \\
bias4 & 1.15 & 1.20 & 0.93 \\
bias5 & 1.36 & 1.22 & 0.92 \\
bias6 & 0.17 & 0.55 & 1.22 \\
\hline
\end{tabular}
}

\label{tab:transistor_sizing}
\end{table}

\begin{figure}[t]
    \centering
    \resizebox{0.49\textwidth}{!}{%
        \includegraphics{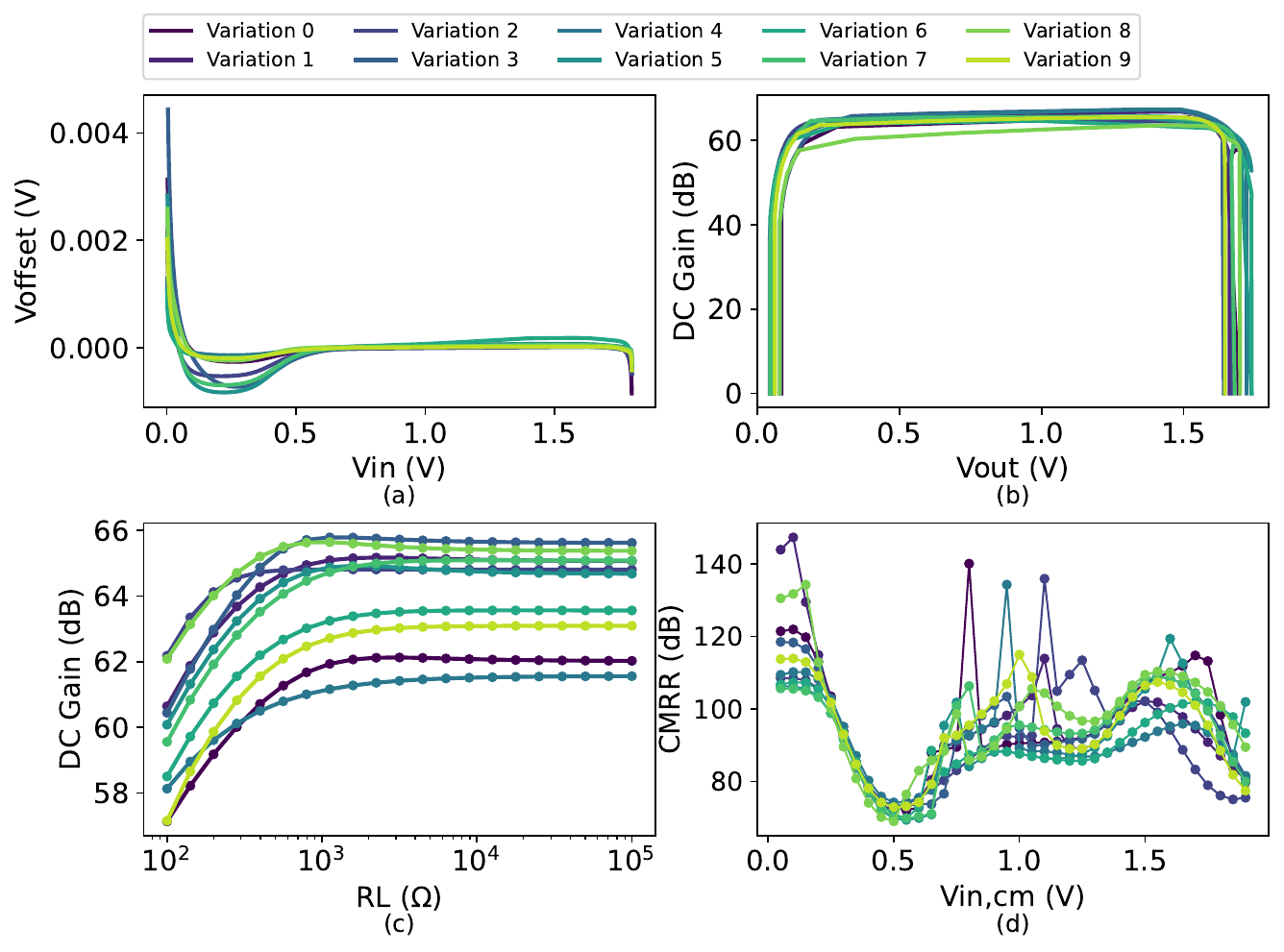}
    }
\caption{Variation results for G1-5. (a) Input Offset Voltage vs. Common-Mode Input. (b) DC Gain vs. Output Voltage (c) DC Gain vs. Load Resistor. (d) CMRR vs. Common Mode Input. Random variations following a Gaussian distribution with a mean ($\mu$) of 0 and a standard deviation ($\sigma$) of 0.1 are added to bias voltages and $\sigma$ of 0.01 for transistor size.}
\label{fig:opamp}
\end{figure}

To further verify the applicability of the optimized circuit, we conducted various tests, including DC, transient and parametric variations on circuit G1-5. These variations include transistor sizes and bias voltages. The results from these tests are shown in Fig.~\ref{fig:opamp}, which confirm design robustness and suitability for practical use under varying conditions.

\section{Discussion}
This work found that the proposed LLM-based agent can effectively propose circuit optimization strategies based on previous and current performance to balance various metrics, demonstrating the potential of LLMs in the circuit sizing process. However, some limitations remain. 

Firstly, the circuit under test relies on bias voltages due to the complexity of biasing circuits, which needs to be replaced with a current mirror network. The overall area of transistors is excluded in this study as the focus remains on performance optimization. Future work should integrate area and layout considerations for a holistic PPA trade-off.

Secondly, the open-loop and closed-loop configurations of the op-amp are tested with a common-mode voltage at half-rail. While the design targets rail-to-rail output with a 1.75~V output range, the input common-mode range is not targeted and optimized, leading to variations in CMRR and DC gain across the input range. Future work should simulate dynamic parameters across the full range for true rail-to-rail input/output performance. Additionally, high variations can be seen in the variation test, especially in Fig.~\ref{fig:opamp} (c) and (d). A true Monte Carlo simulations with accurate device models are needed for the agent to ensure the design robustness. Moreover, optimization was limited to 25 iterations and 5 trials due to rate limits, potentially restricting performance insights and affecting the reliability of the results. However, more iterations don’t ensure convergence, as LLMs may oscillate between solutions. A feedback mechanism should be introduced to prompt diverse solutions and improve reliability.

\section{Conclusion}
This work utilizes LLMs to build an AI Agent for the AMS circuit sizing process. By employing function-calling techniques, the LLMs are integrated with the external simulator Ngspice and data analysis functions, Additionally, using prompt engineering strategies, such as CoT and in-context learning, the agent leverages contextual information from previous iterations to guide the LLMs in sizing, leading to a reduced the number of iterations. As a result, the proposed agent optimized seven small-sized circuits, including a twenty-transistor rail-to-rail opamp. This agent demonstrates the potential for optimizing more complex circuits and systems in the future.

The source code for this project is available on \blackout{https://github.com/eelab-dev}

\section*{Acknowledgements}

\blackout{The authors thank EDINA and ISG@University of Edinburgh for their support in accessing OpenAI services.}

%\begin{thebibliography}{00}
\bibliographystyle{IEEEtran}
\bibliography{IEEEabrv,bib}
%\end{thebibliography}
\vspace{12pt}
\end{document}